\documentclass[runningheads]{llncs}
\usepackage[T1]{fontenc}
\usepackage{xcolor}
\usepackage{graphicx,verbatim}
\usepackage{amsmath,amssymb}
\usepackage{algorithm}
\usepackage{algpseudocode}
\usepackage{booktabs}
\usepackage{bm}
\usepackage[normalem]{ulem}
\usepackage{marvosym}

\newcommand{\ms}[2]{#1{\scriptstyle\pm #2}}
\newcommand{\msb}[2]{\bm{#1}{\scriptstyle\pm \bm{#2}}}

\begin{document}
\title{DreamReg: Belief-Driven World Model for 2D--3D Ultrasound Registration}
\begin{comment}  %% Removed for anonymized MICCAI submission
\author{First Author\inst{1}\orcidID{0000-1111-2222-3333} \and
Second Author\inst{2,3}\orcidID{1111-2222-3333-4444} \and
Third Author\inst{3}\orcidID{2222--3333-4444-5555}}
%
\authorrunning{F. Author et al.}
\institute{Princeton University, Princeton NJ 08544, USA \and
Springer Heidelberg, Tiergartenstr. 17, 69121 Heidelberg, Germany
\email{lncs@springer.com}\\
\url{http://www.springer.com/gp/computer-science/lncs} \and
ABC Institute, Rupert-Karls-University Heidelberg, Heidelberg, Germany\\
\email{\{abc,lncs\}@uni-heidelberg.de}}

\end{comment}

\author{Luoyao Kang\inst{1}\index{Kang, Luoyao} \and 
    Yuelin Zhang\inst{1}\index{Zhang, Yuelin} \and 
    Jiwei Shan\inst{1}\index{Shan, Jiwei} \and 
    Haifan Gong\inst{3}\index{Gong, Haifan} \and 
    Qingpeng Ding\inst{1}\index{Ding, Qingpeng} \and
    Shing Shin Cheng\inst{1,2}$^{\textrm{\Letter}}$\index{Cheng, Shing Shin}
    } 
\authorrunning{L. Kang \textit{et al.}}
\institute{
Department of Mechanical and Automation Engineering and T Stone Robotics Institute, The Chinese University of Hong Kong, Hong Kong SAR, China
\and
Shun Hing Institute of Advanced Engineering and Multi-scale Medical Robotics Center, Hong Kong SAR, China \\
\email{sscheng@cuhk.edu.hk}
\and
Perelman School of Medicine, University of Pennsylvania, PA, USA
    }
  
\maketitle              % typeset the header of the contribution
\begin{abstract}
Ultrasound (US) is widely used for surgical navigation, yet real‑time registration between intraoperative 2D slices and preoperative 3D volumes remains challenging due to partial observability, speckle noise, and the action‑dependent US acquisition. Existing methods are one‑shot or short‑horizon, making it hard for them to gather evidence over time or capture how surgeons adjust probe motion based on on‑screen feedback.
We propose DreamReg, a belief-driven world-model framework that formulates 2D--3D registration as belief updating over rigid transformations. DreamReg maintains a latent belief state that summarizes past observations and poses information, and continuously refines the transformation through learned dynamics as new slices arrive. 
During training, DreamReg is exposed to probe-motion trajectories that mimic clinical scanning behavior and learns to update its belief by conditioning pose refinement on the current US observation. During inference, DreamReg refines registration via internal imagination: it rolls out the learned world model to simulate candidate probe motions and their predicted observations, and integrates these imagined outcomes to converge to an accurate rigid transformation. 
Experiments on CAMUS and $\mu$-RegPro datasets demonstrate improved robustness and competitive registration accuracy for real-time guidance compared with state-of-the-art methods.
Project page: \url{https://github.com/kangluoyao/DreamReg}.
% Authors must provide keywords and are not allowed to remove this Keyword section.

\end{abstract}
\section{Introduction}

Ultrasound (US) is widely used for surgical navigation due to its real-time and safe imaging capabilities~\cite{gong2023thyroid,pavone2024ultrasound,smit2022ultrasound}. 
In many workflows, a preoperative 3D US volume provides global context, while intraoperative guidance relies on streaming 2D slices~\cite{ferrante2017slice,gillies2017real,hu2012mr}, making real-time 2D--3D registration essential for navigation. 
However, registration is fundamentally challenging: 2D slices offer partial, speckle-corrupted observations, and subsequent views depend on probe motion, leading to strong action-observation coupling under partial observability~\cite{liu2019deep}.
% {\color{blue}this sentence is a bit strange with two `observations' used. May want to separate it into two sentences.}

Learning-based approaches for ultrasound registration can be broadly categorized into two paradigms.
The first learns 2D/3D features and recovers rigid transformations via geometric fitting~\cite{fischler1981random}, leveraging CNN descriptors to establish correspondences before solving for 6-DoF pose~\cite{brandstatter2024rigid,markova2022global}.
% Although interpretable and robust to outliers, these pipelines rely on repeatable local structures, which are unreliable in speckle-rich, view-dependent US images.
The second paradigm performs end-to-end pose regression~\cite{guo2021end,lei2024epicardium,wang2025eureg,yeung2021learning} models, 
% such as FVR-Net~\cite{guo2021end} performs end-to-end pose regression with differentiable resampling.
% EUReg~\cite{wang2025eureg} enhances crossing dimension interaction via flow modeling for improved efficiency.
% CUReg~\cite{lei2024epicardium} introduces anatomy-aware prompts and inter-frame regularization to enhance the US feature.
% Despite these advances, most methods assume passively observed slices and estimate pose in a one-shot manner.
but typically treat slices as passively observed and operate in a one-shot or short‑horizon manner.
In practice, clinicians actively adjust probe pose based on visual feedback to disambiguate anatomy~\cite{bahner2016language,mulder2023unravelling,weld2025identifying}.
Ignoring this coupling often leads to brittle performance under sparse views.
Recent advances in world model provides a framework for decision-making under partial observability by learning latent dynamics that couple actions and observations~\cite{ha2018world,hafnerdream,hafner2020mastering}.
% {\color{blue}Dont start a sentence with `And'.}
% And
It demonstrated strong performance in robotics and control by enabling internal simulation and imagination-based planning without exhaustive real-world interaction~\cite{lu2025gwm,wu2023daydreamer}.
In medical imaging, however, world-model formulations remain largely unexplored~\cite{mu2026ehrworld,yang2025medical}, and existing registration approaches rarely model how probe motion influences future observations through learned latent dynamics.
% {\color{blue}this sentence is a bit strange with two `observations' used. May want to separate it into two sentences.}

% {\color{blue}Overall, there should be significantly more review of the state-of-the-art (SOTA) in the introduction. We want to know what the SOTA works are in this field to allow us to know the missing gap in the field. What are the different approaches to do US 2d-3d registration? Maybe you can categorize them into 2-3 main categories (traditional and AI based? Maybe there are different AI approaches too? conventional learning-based one-shot? other more recent AI approaches?)? What are their superiority and limitations?}

Motivated by this gap, we propose \textbf{DreamReg}, a belief-driven world-model framework that reformulates 2D--3D US registration as sequential belief updating over rigid transformations.
Rather than predicting pose from an isolated slice, DreamReg maintains a latent belief state representing the current alignment hypothesis under partial observability~\cite{ha2018world,hafnerdream}.
This belief evolves through learned action-conditioned dynamics, coupling probe motion and observation transitions in latent space.
During training, monitored probe maneuvers supervise belief transitions, yielding a data-driven prior over clinically plausible scanning dynamics.
During inference, DreamReg performs imagination-based rollouts: the learned world model simulates candidate probe motions and predicted observations to iteratively refine belief, without dense correspondences or heuristic exploration.
By belief updating and action-conditioned modeling, DreamReg departs from conventional one-shot regression and improves robustness in challenging intraoperative scenarios.
% Extensive experiments demonstrate competitive accuracy compared with strong end-to-end baselines and state-of-the-art 2D–3D US registration methods.

% \paragraph{Contributions.}
\noindent\textbf{Contributions.}
(1) Real-time 2D--3D ultrasound registration is reformulated as a belief-driven sequential decision problem under partial observability.
(2) An action-conditioned world model is proposed to explicitly capture the coupling between probe motion and observation in a latent space for robust pose refinement.  
(3) A belief rollout mechanism is introduced to enable model-based imagination during inference.
% decoupling pose refinement from external observation feedback.
(4) Comprehensive evaluations are conducted on CAMUS~\cite{leclerc2019deep} and $\mu$-RegPro~\cite{zachary_m_c_baum_2023_7861715}, where consistent improvements in both geometric accuracy and image similarity are demonstrated over the state-of-the-art baselines.

\section{Methodology}
\label{sec:method}

\subsection{Problem Formulation}
\begin{figure}[t]
    \centering
    \includegraphics[width=0.9\linewidth]{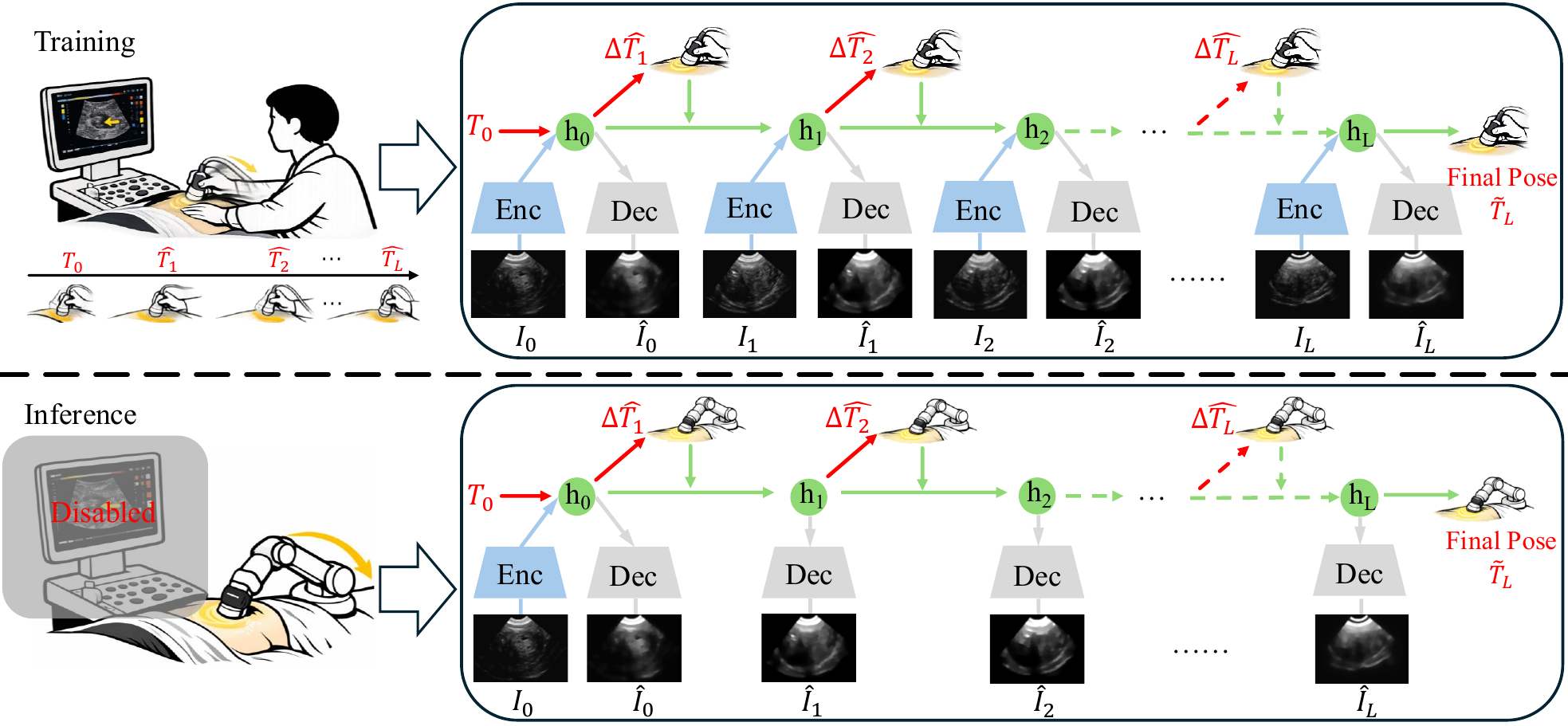}
    \caption{\textbf{Belief-driven world-model registration in training and inference.}
    \emph{Training (top):} Mimicking a clinician with screen feedback, the model iteratively refines the pose; 
    % by rolling out a sequence of actions $\{\Delta\hat T_t\}_{t=1}^{L}$; 
    % each observed slice $o_t$ updates the belief state $h_t$, and the model generates a corresponding reconstruction $\hat{o}_t$.
    \emph{Inference (bottom):} Without observations, the model performs closed-loop refinement by imagination.
    % , using $\hat{o}_t$ generated by the decoder to update $h_t$.
    }
    \label{fig:overview}
\end{figure}

% \sout{We study rigid 2D--3D ultrasound (US) registration for image-guided navigation.}
Given a preoperative 3D US volume
$V \in \mathbb{R}^{1 \times D \times H \times W}$ and an intraoperative target 2D US slice
$I_g \in \mathbb{R}^{1 \times H \times W}$, the objective is to estimate a 6-DoF rigid US probe pose
$T^\star \in \mathbb{R}^{6}$ such that rendering a slice from the volume at $T^\star$
matches $I_g$.
A differentiable physics-based slice renderer is given by $I(T_{t}) = \mathcal{R}(V, T_{t}) \in \mathbb{R}^{1 \times H \times W}$ with a noise-perturbed expert pose $T_t$ and a sampling operator $R$.
Instead of directly regressing $T^\star$ in a one-shot manner, we formulate registration as an iterative decision process, as shown in Fig.\ref{fig:overview}:
starting from an initial pose $T_0$, at each step $t$ the agent predicts an incremental motion
$\Delta\hat T_t \in \mathbb{R}^{6}$ and updates the probe-pose estimate as $\tilde{T}_{t+1} = \tilde{T}_t + \Delta\hat T_t$. Here, $\tilde{T}_t$ denotes the agent's current estimate of the probe pose at step $t$.

\begin{figure}[t]
  \centering
  \includegraphics[width=0.8\linewidth]{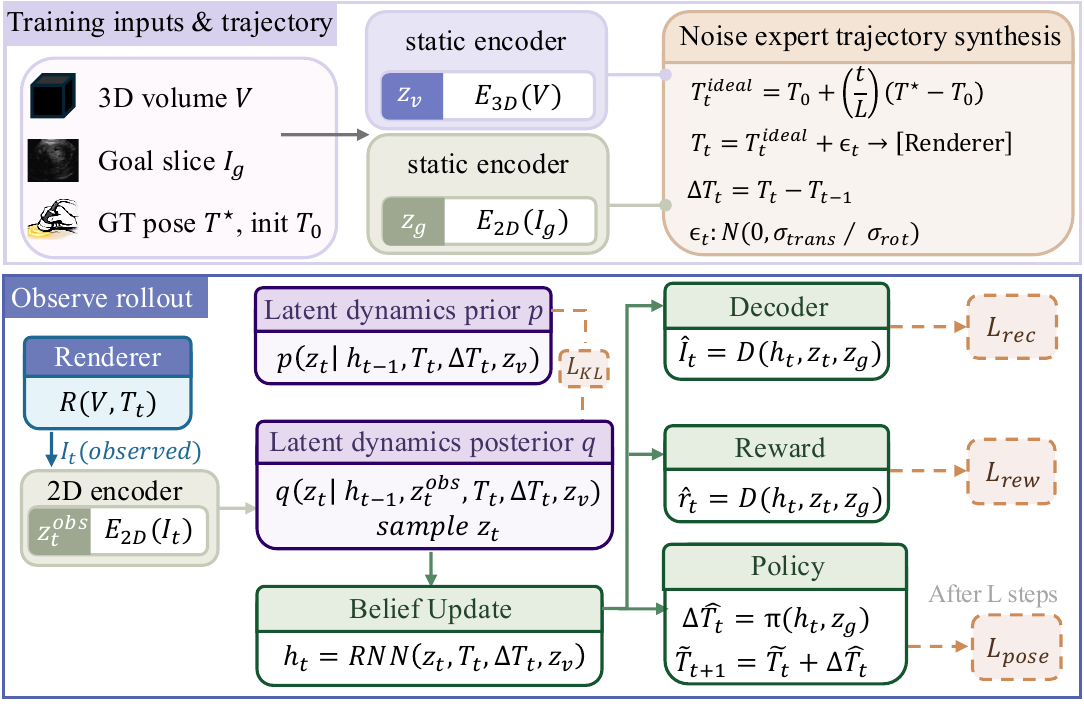}
  % \caption{\textbf{Training-time rollout and world-model learning.}
  % Given $(V, I_g, T_0, T^{\ast})$, 3D/2D encoders compute $(z_v, z_g)$.
  % A noise-perturbed expert pose $T_t$ is rendered to obtain $I_t=R(V,T_t)$ and $z_t^{\mathrm{obs}}$ for forced rollouts.
  % The world model updates belief $h_t$ via posterior inference regularized by a learned prior (KL divergence), and trains a decoder/policy to predict $(\hat I_t,\hat r_t)$ and pose increments $\Delta\hat T_t$.
  % Losses are accumulated over $t=1,\ldots,L$.
  % }
  \caption{\textbf{Training-time rollout}.
    % \emph{Top (purple)}: we construct a noise-perturbed expert trajectory from $(T_0\!\rightarrow\!T^{\ast})$ and render an observation rollout $\{I_t=R(V,T_t)\}$, while static encoders extract $(z_v,z_g)$.
    % \emph{Bottom (blue)}: DreamReg performs an observe rollout for world-model learning: rendered slices are encoded to $z_t^{\mathrm{obs}}$ and used by the \emph{posterior} (purple) to sample $z_t$, regularized toward the \emph{prior} via KL divergence; the belief state $h_t$ is then updated recurrently (green).
    % Conditioned on $(h_t,z_t,z_g)$, decoder/reward heads predict $(\hat I_t,\hat r_t)$ and the policy outputs pose increments $\Delta\hat T_t$ (with $z_g$ used only for task conditioning).
    % Losses are accumulated over $t=1,\ldots,L$.
    A noise-perturbed expert trajectory is rendered to produce observations.
    The posterior infers latent states conditioned on observations and is regularized by the prior via the KL divergence.
    The belief state is updated recurrently and used for slice reconstruction, reward prediction, and pose refinement.
    }
  \label{fig:training_pipeline}
\end{figure}

\subsection{Belief-Driven World Model for Registration}
\noindent\textbf{Model overview.}
As illustrated in Fig.~\ref{fig:overview}, training (top) performs an observe rollout where rendered slices are encoded to update the belief state via posterior inference, whereas inference (bottom) performs a closed-loop imagination rollout using only the learned prior without new observations.
Our model, \texttt{DreamReg}, follows a Dreamer-style~\cite{hafnerdream,hafner2020mastering} latent world model for iterative pose refinement.
At each step $t$, the model maintains a latent belief state composed of: 
(i) a \emph{stochastic latent variable} $z_t$, which represents the latent encoding of the current slice conditioned on pose and belief (see the prior/posterior blocks in Fig.~\ref{fig:training_pipeline}). 
% $z_t$ explicitly models pose uncertainty under partial observability.
$z_t$ captures a stochastic representation of the latent observation state under partial observability.
And (ii) a \emph{belief update} $h_t$, which aggregates temporal information through recurrent updates (Fig.~\ref{fig:overview}, green nodes).
The latent transition is modeled by a learned prior $p_\theta(z_t \mid \cdot)$, while a posterior $q_\phi(z_t \mid \cdot)$ incorporates the rendered observation during training.
The sampled $z_t$ updates the belief state $h_t$, and conditions
slice reconstruction, reward prediction, and policy learning
(Fig.~\ref{fig:training_pipeline}).
% At each step $t$, DreamReg maintains a latent belief state consisting of a stochastic variable $z_t$ and a belief state 
% $h_t$ (Fig.~\ref{fig:training_pipeline}). 
% The prior $p_\theta(z_t \mid \cdot)$ models action-conditioned latent transitions, 
% while the posterior $q_\phi(z_t \mid \cdot)$ incorporates rendered observations during training. 
% The sampled $z_t$ updates the belief state $h_t$ and conditions slice reconstruction, reward prediction, and policy learning.

% {\color{blue}Reference Fig. 2 to help readers understand better your method and workflow. Perhaps a }

% \paragraph{Encoders for static context and goal.}
\noindent\textbf{Encoders for static context and goal.}
As shown in Fig.~\ref{fig:training_pipeline}, 
the 3D volume $V$ and target slice $I_g$ are encoded by 3D and 2D CNNs, yielding embeddings $z_v$ and $z_g$.
% Concretely, a 3D CNN encoder $E_{3D}$ extracts volumetric features $f_v = E_{3D}(V)$, while a 2D CNN encoder $E_{2D}$ produces slice features $f_g = E_{2D}(I_g)$. 
% Both features are then projected to a common latent dimension $d_z$ via linear layers:
% $z_v = P_v(f_v), z_g = P_s(f_g).$
% These modules correspond to the static encoder blocks in Fig.~\ref{fig:training_pipeline}.

\noindent\textbf{Latent dynamics with prior/posterior.}
At training time, the current slice is rendered as $I_t=\mathcal{R}(V,T_t)$ along a known pose trajectory.
% We encode $I_t$ into an observation embedding $z^{\text{obs}}_t = P_s(E_{2D}(I_t))$.
The slice $I_t$ is encoded into an observation embedding $z^{\text{obs}}_t = P_s(E_{2D}(I_t))$.
The world model defines a Gaussian prior and posterior over $z_t$.
The prior predicts the next latent state without observing the current slice,
while the posterior incorporates the rendered observation during training
(see purple blocks in Fig.~\ref{fig:training_pipeline}):
\begin{align}
p_\theta(z_t \mid h_{t-1}, \Delta T_t, T_t, z_v)
&= \mathcal{N}\!\big(\mu^p_t, \mathrm{diag}(\sigma^{p2}_t)\big), \\
q_\phi(z_t \mid h_{t-1}, \Delta T_t, T_t, z_v, z^{\text{obs}}_t)
&= \mathcal{N}\!\big(\mu^q_t, \mathrm{diag}(\sigma^{q2}_t)\big),
\end{align}
where $(\mu,\log\sigma^2)$ are produced by two MLPs.
Sampling is performed using the reparameterization trick:
% \begin{equation}
$z_t = \mu^q_t + \sigma^q_t \odot \epsilon, \epsilon\sim\mathcal{N}(0,I).$
% \end{equation}

\noindent\textbf{Belief update.}
The belief state is updated by a recurrent module applied to a projected input:
\begin{equation}
u_t = W_u [\, z_t;\, \Delta T_t;\, T_t;\, z_v \,], \quad
h_t = \mathrm{RNN}(u_t, h_{t-1}),
\end{equation}
where $[\,\cdot\,;\,\cdot\,]$ denotes concatenation and $\mathrm{RNN}$ is a recurrent transformer block~\cite{bulatov2022recurrent}.
Importantly, we do \emph{not} feed the goal embedding $z_g$ into the latent dynamics update to prevent shortcut learning, where the model directly learns the final pose from the goal embedding.
Instead, $z_g$ only functions as a \emph{task condition} for decoding and reward.
This belief update accumulates information across time steps.

\noindent\textbf{Imagination decoder and reward model.}
Given $(h_t,z_t,z_g)$, we decode an imagined slice $\hat I_t$ and predict a scalar reward vector $\hat r_t$ (Fig.~\ref{fig:training_pipeline}, Decoder and Reward head):
\begin{equation}
\hat I_t = D_\psi([h_t;z_t;z_g]),\qquad 
\hat r_t = R_\psi([h_t;z_t;z_g]),
\end{equation}
where $D_\psi$ reconstructs the slice from the latent belief,  and $R_\psi$ outputs two values measuring translational and rotational progress.
The decoder provides reconstruction supervision for world-model learning, while the reward head supplies dense progress signals for action supervision.

\noindent\textbf{Policy for pose refinement.}
At the inference phase, real slices are no longer needed while the pose is purely refined by imagination (Fig.~\ref{fig:overview}, bottom).
The policy predicts pose increments from the current belief and goal embedding: $\Delta \hat T_t = \pi_\omega([h_t; z_g]).$
The belief is then updated using the learned prior:
\begin{equation}
z_t \sim p_\theta(z_t \mid h_{t-1},\Delta \hat T_t,T_t,z_v),\quad
h_t = \mathrm{RNN}(W_u[ z_t;\Delta \hat T_t;T_t;z_v], h_{t-1}).
\end{equation}

Importantly, the prior models how the latent state evolves \emph{given} an action, whereas the policy predicts \emph{which} action should be taken.
The two components therefore play complementary roles: the prior learns latent transition dynamics, while the policy performs action selection from the belief state.
Without the policy head, the model would not be able to generate pose updates during inference, since the prior requires an action input but does not predict one.

\subsection{Training via Observe Rollout}
\label{sec:training}
Training follows the observe rollout pipeline shown in Fig.~\ref{fig:training_pipeline}, which uses supervised pose labels $T^\star$ for each pair $(V, I_g)$. 
We synthesize a length-$L$ trajectory from $T_0$ to $T^\star$
and inject noise to emulate off-trajectory states.
% , which is crucial for learning stable dynamics and robust policies.

% \paragraph{Trajectory synthesis.}
\noindent\textbf{Trajectory synthesis.}
Let $L$ be the number of world-model steps. We define an ``ideal'' linear path
$T^{\text{ideal}}_{t}=T_0 + \tfrac{t}{L}(T^\star-T_0)$ for $t=1\ldots L$, and then add Gaussian noise $\eta_t\sim\mathcal{N}(0,\sigma^2 I)$ to obtain $T_t$.
Actions are defined consistently as differences between consecutive noisy poses $\Delta T_t = T_t - T_{t-1}$.
% \begin{equation}
% \Delta T_t = T_t - T_{t-1}.
% \end{equation}
% This ensures the world model learns transition behavior conditioned on the \emph{current} pose.

\noindent\textbf{Observe rollout.}
Given $(V, I_g, \{T_t\}_{t=1}^L, \{\Delta T_t\}_{t=1}^L)$, we run an observe rollout:
at each step, we render $I_t=\mathcal{R}(V,T_t)$, infer the posterior $q_\phi(z_t|\cdot)$, update belief $h_t$, decode $\hat I_t$, predict $\hat r_t$, and accumulate the KL divergence.

\noindent\textbf{World-model learning.}
We train DreamReg with a sum of rollout-averaged losses: $\mathcal{L}=
\mathcal{L}_{\mathrm{KL}}+
\mathcal{L}_{\mathrm{rec}}+
\mathcal{L}_{\mathrm{rew}}+
\mathcal{L}_{\mathrm{pose}}$.
Latent dynamics are regularized by minimizing the KL divergence between the posterior and the prior, while the decoder reconstructs the rendered slices. Formally, we define
\begin{equation}
\mathcal{L}_{\mathrm{KL}}=\mathbb{E}_{t}\!\left[D_{\mathrm{KL}}\!\left(q_\phi(z_t|\cdot)\|p_\theta(z_t|\cdot)\right)\right],\quad
\mathcal{L}_{\mathrm{rec}}=\mathbb{E}_{t}\!\left[\|\hat I_t-I_t\|_1+\mathcal{L}_{\mathrm{SSIM}}(\hat I_t,I_t)\right],
\end{equation}
where $\mathbb{E}_{t}$ denotes the average over the $L$ time steps ($t=1,\dots,L$).

\noindent\textbf{Progress reward and pose supervision.}
We supervise the reward head by aligning predicted actions with the direction to the ground-truth pose.
The reward head provides supervision on directional progress.
% stabilizing policy learning beyond the terminal pose loss.
Let $\tilde T_t$ be the policy-updated pose and $d_t=T^\star-\tilde T_t$, we define:
\begin{equation}
    r_t=[\cos(\hat\Delta T_t^{1:3},d_t^{1:3}),\,\cos(\hat\Delta T_t^{4:6},d_t^{4:6})],\quad
    \mathcal{L}_{\mathrm{rew}}=\frac{1}{L}\sum_{t=1}^{L}\|\hat r_t-r_t\|_2^2
\end{equation}
% \sout{A terminal SmoothL1 loss enforces final pose accuracy:}
Pose prediction is supervised by a Smooth$\mathcal{L}_1$ loss:
\begin{equation}
    \mathcal{L}_{\mathrm{pose}}=\mathrm{SmoothL1}(\tilde T_L^{1:3},T^{\star\,1:3})+\mathrm{SmoothL1}(\tilde T_L^{4:6},T^{\star\,4:6}).
\end{equation}

\subsection{Inference via Imagination Rollout}
During inference, we perform the imagination rollout shown in Fig.~\ref{fig:overview} (bottom).
Given $(V,I_g)$, we compute $z_v$ and $z_g$ once and initialize $T_0=\mathbf{0}, h_0=\mathbf{0}$.
For $L$ steps:
(i) predict action $\Delta \hat{T}_t=\pi_\omega([h_t;z_g])$,
(ii) update pose $\tilde{T}_{t+1}=\tilde{T}_t+\,\Delta T_t$ with angle wrapping to $(-\pi,\pi]$,
(iii) sample $z_t \sim p_\theta(\cdot)$ and update belief via the prior.
This closed-loop imagination refinement enables pose estimation without real-time image feedback.
The final pose $\tilde{T}_L$ is returned.

\begin{table}[t]
\centering
\setlength{\tabcolsep}{1.1pt}
\renewcommand{\arraystretch}{1.0}
\small
\caption{\textbf{State-of-the-art comparison on CAMUS and $\mu$-RegPro (mean${\scriptstyle\pm}$std).}
Arrows indicate whether lower ($\downarrow$) or higher ($\uparrow$) is better.
% Values are mean${\scriptstyle\pm}$std. FPS is measured on the same hardware and inference setting.
}
\label{tab:sota_combined_fps}
\begin{tabular}{@{}lccccccc@{}}
\toprule
\multicolumn{8}{c}{\textbf{CAMUS}} \\
\midrule
Method & DisErr$\downarrow$ & I-NCC$\uparrow$ & SSIM$\uparrow$
& TransErr$\downarrow$ & RotErr$\downarrow$ & P-NCC$\uparrow$ & FPS$\uparrow$ \\
\midrule
CUReg~\cite{lei2024epicardium}
& $\ms{9.77}{0.03}$ & $\ms{79.72}{0.15}$ & $\ms{38.48}{0.65}$
& $\ms{7.60}{0.03}$ & $\ms{7.75}{0.02}$ & $\ms{53.10}{0.75}$ & 37 \\
EUReg~\cite{wang2025eureg}
& $\ms{8.40}{0.33}$ & $\ms{86.69}{1.54}$ & $\ms{44.33}{2.20}$
& $\ms{5.59}{0.32}$ & $\ms{7.98}{0.07}$ & $\ms{64.55}{1.19}$ & \textbf{189} \\
FVRNet~\cite{guo2021end}
& $\ms{9.38}{1.14}$ & $\ms{82.23}{5.01}$ & $\ms{40.08}{4.15}$
& $\ms{6.72}{1.14}$ & $\ms{8.10}{0.22}$ & $\ms{58.49}{7.58}$ & 55 \\
DreamReg
& $\msb{7.16}{0.13}$ & $\msb{90.45}{0.14}$ & $\msb{51.87}{0.35}$
& $\msb{4.77}{0.07}$ & $\msb{7.08}{0.15}$ & $\msb{72.12}{1.62}$ & 40 \\
\midrule
\multicolumn{8}{c}{\textbf{$\mu$-RegPro}} \\
\midrule
Method & DisErr$\downarrow$ & I-NCC$\uparrow$ & SSIM$\uparrow$
& TransErr$\downarrow$ & RotErr$\downarrow$ & P-NCC$\uparrow$ & FPS$\uparrow$ \\
\midrule
CUReg~\cite{lei2024epicardium}
& $\ms{12.43}{0.48}$ & $\ms{55.19}{6.18}$ & $\ms{22.27}{2.60}$
& $\ms{8.53}{0.72}$ & $\ms{9.30}{0.03}$ & $\ms{22.37}{15.38}$ & 36 \\
EUReg~\cite{wang2025eureg}
& $\ms{11.89}{0.06}$ & $\ms{63.03}{0.57}$ & $\ms{27.61}{0.43}$
& $\ms{7.74}{0.12}$ & $\ms{9.29}{0.01}$ & $\ms{36.26}{2.10}$ & \textbf{202} \\
FVRNet~\cite{guo2021end}
& $\ms{11.83}{0.66}$ & $\ms{58.99}{6.63}$ & $\ms{24.67}{3.97}$
& $\ms{7.71}{0.95}$ & $\ms{9.31}{0.02}$ & $\ms{34.85}{10.90}$ & 51 \\
DreamReg
& $\msb{11.06}{0.24}$ & $\msb{66.68}{2.24}$ & $\msb{30.99}{1.95}$
& $\msb{6.86}{0.36}$ & $\msb{9.19}{0.02}$ & $\msb{44.39}{3.34}$ & 56 \\
\bottomrule
\end{tabular}
\end{table}

\section{Experiments and Results}

\subsection{Datasets and Implementation Details}

We evaluate DreamReg on two public 3D ultrasound datasets: CAMUS~\cite{leclerc2019deep} and the ultrasound subset of $\mu$-RegPro~\cite{zachary_m_c_baum_2023_7861715}.
For CAMUS, we follow the official split. The dataset contains 1000 cardiac US volumes. We uniformly sample 4 slices per case and resample each 3D volume to $(32,192,192)$.
For $\mu$-RegPro, we use a case-level split with a 7:1:2 ratio for training/validation/testing, which contains 73 prostate volumes in total. 
We sample 8 slices per case and resample volumes to $(64,64,64)$.
During training, the initial 6-DoF pose perturbations are randomly sampled within $\pm 10$ (mm for translation and degrees for rotation) to simulate realistic misalignment, consistent with prior frame-to-volume registration settings~\cite{lei2024epicardium,wang2025eureg}.
Following EUReg~\cite{wang2025eureg}, training uses random sampling on-the-fly, while for evaluation, we pre-sample a fixed test set during preprocessing to ensure reproducibility (400 slices for CAMUS and 120 for $\mu$-RegPro). 
% The resulting test sets contain 400 slices for CAMUS and 120 slices for $\mu$-RegPro. 
% All reported numbers are mean$\pm$std over the corresponding test set.

All experiments are conducted on a single NVIDIA A800 GPU with a batch size of 64 for 100 training epochs.
Optimization is performed using AdamW~\cite{loshchilov2017decoupled} with a two-stage learning rate schedule.
The initial learning rate is set to $1\times10^{-4}$ for CAMUS and $1\times10^{-5}$ for $\mu$-RegPro.
% A two-stage learning rate schedule is adopted: a StepLR scheduler is first applied for warm-up, followed by a ReduceLROnPlateau scheduler driven by validation performance.
The number of world-model rollout steps $L$ is set to 7 for CAMUS and 5 for $\mu$-RegPro.
For evaluation, we report seven metrics covering geometric accuracy, image alignment, parameter consistency, and efficiency~\cite{lei2024epicardium,wang2025eureg}.
Specifically, we use DisErr (mm), TransErr (mm), and RotErr ($^\circ$) for pose accuracy; I-NCC (\%) and SSIM (\%) for image-level alignment~\cite{rao2014application,wang2004image}; P-NCC (\%) for parameter consistency; and frames per second (FPS) for runtime efficiency.

\begin{figure*}[t]
    \centering
    \includegraphics[width=0.95\textwidth]{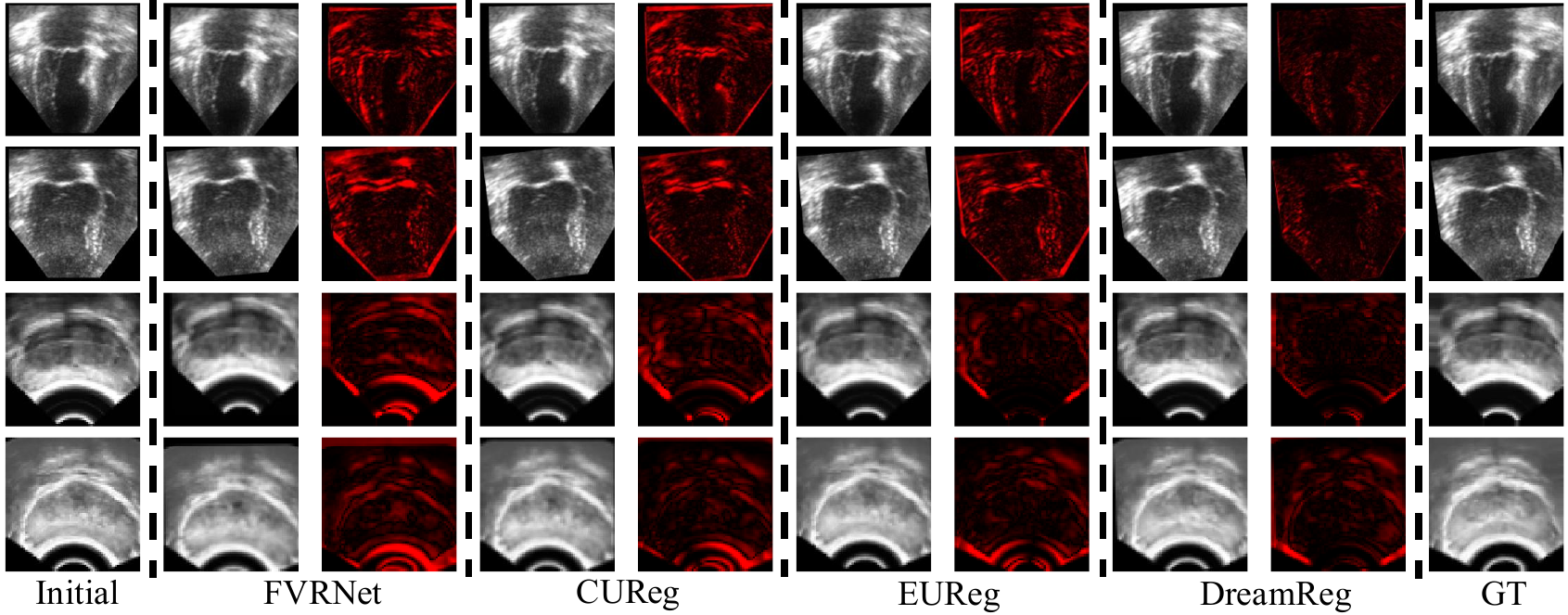}
    \caption{
    \textbf{Qualitative comparison on registration results of different methods.}
    Top two rows: CAMUS; bottom two rows: $\mu$-RegPro. 
    Red overlays indicate intensity differences between the target slice and the resampled slice.
    % {\color{blue}better to have 1 more example (another row).}
    }
    \label{fig:qualitative}
\end{figure*}

\subsection{Comparison with the State-of-the-art Methods}
% {\color{blue}I think results analysis is not strong enough. It is also quite repetitive and wordy. e.g., it restates numbers multiple times across text and tables, and key messages are written in long sentences. I would suggest to focus on providing a big summary of the performance comparison and hihglight some key difference/outperformance and explain the reason and connect this outperformance to clinical needs to highlight its clinical signifincance. Use AI tools to help with addressing this problem.}

We compare DreamReg with the SOTA 2D--3D registration baselines,
including CUReg~\cite{lei2024epicardium}, EUReg~\cite{wang2025eureg}, and FVRNet~\cite{guo2021end},
on CAMUS~\cite{leclerc2019deep} and $\mu$-RegPro~\cite{zachary_m_c_baum_2023_7861715}.

\noindent\textbf{Overall performance across datasets.}
Across both CAMUS and $\mu$-RegPro (Table~\ref{tab:sota_combined_fps}), DreamReg consistently achieves the best overall performance in both geometric accuracy and image similarity metrics, demonstrating the robustness of belief-driven latent modeling over one-shot regression approaches. 
On both datasets, DreamReg yields the lowest distance and translation errors while simultaneously achieving the highest I-NCC, SSIM, and P-NCC scores. 
Notably, the improvements are particularly pronounced in similarity-based metrics, indicating enhanced structural alignment rather than merely reduced pose discrepancies. 
Compared with the strongest baseline (EUReg), DreamReg consistently increases structural similarity across datasets, suggesting that iterative belief updates and latent representation modeling lead to more anatomically coherent registrations. 
The consistent gains across multiple datasets indicate that explicitly modeling the interaction between action and observation produces more stable and clinically meaningful alignment.

\noindent\textbf{Analysis of runtime and practical considerations.}
% In addition to accuracy, we report inference speed (FPS) in Table~\ref{tab:sota_combined_fps}.
% While EUReg achieves higher raw throughput (189–202 FPS), DreamReg runs at 40–56 FPS, which still satisfies real-time requirements (FPS $>$ 30).
% Importantly, ultrasound-guided procedures prioritize low-latency and stable feedback over ultra-high frame rates~\cite{wei2015real,giangrossi2022requirements,hidalgo2025evaluating}.
% In typical clinical workflows~\cite{wei2015real}, probe motion and anatomical response occur at human-controlled speeds, where consistent alignment and robustness are more critical than excessive rendering frequency.
% The moderate computational overhead of DreamReg stems from iterative belief updates and latent dynamics modeling, which explicitly capture action–observation coupling.
% This design trades marginal throughput for substantially improved geometric and structural accuracy.
% Given that 40+ FPS already exceeds the perceptual threshold for smooth visual feedback, the additional speed margin provided by faster baselines does not translate into meaningful clinical benefit.
% Instead, improved reliability and structural consistency are more valuable for safety-critical navigation scenarios.
In addition to accuracy, we report inference speed (FPS) in Table~\ref{tab:sota_combined_fps}. 
Although EUReg achieves higher throughput (189--202 FPS), DreamReg operates at 40--56 FPS, exceeding real-time requirements (FPS $>$ 30). 
The additional computational cost arises from iterative belief updates and latent dynamics modeling. 
While this reduces raw throughput, it yields improved geometric and structural accuracy. 
Given that 40+ FPS already ensures smooth visual feedback in ultrasound-guided procedures~\cite{giangrossi2022requirements,hidalgo2025evaluating,wei2015real}, 
% \sout{further increases in frame rate provide limited practical benefit compared with improved registration reliability.} 
trading a moderate amount of inference latencies for improved registration reliability is both acceptable and clinically meaningful.

\noindent\textbf{Visualization results.} Fig.~\ref{fig:qualitative} presents qualitative comparisons of different
registration methods. The initial poses exhibit large structural misalignment.
% While all learning-based baselines improve alignment to some extent, residual distortions and boundary inconsistencies remain visible.
DreamReg produces resampled slices that are visually
closer to the ground truth, with reduced error responses in the difference maps,
indicating more accurate and stable pose estimation.

\subsection{Ablation Study}
\begin{table}[t]
\centering
\setlength{\tabcolsep}{3pt}
\renewcommand{\arraystretch}{1.0}
\small
\caption{\textbf{Ablation study on CAMUS (mean${\scriptstyle\pm}$std)}.}
\label{tab:ablation}
\begin{tabular}{lcccccc}
\toprule
Method & DisErr$\downarrow$ & I-NCC$\uparrow$ & SSIM$\uparrow$ & TransErr$\downarrow$ & RotErr$\downarrow$ & P-NCC$\uparrow$ \\
\midrule
only pose & $\ms{7.35}{0.04}$ & $\ms{90.12}{0.23}$ & $\ms{50.81}{0.96}$ & $\ms{4.84}{0.07}$ & $\ms{7.28}{0.06}$ & $\ms{70.85}{0.34}$ \\
w/o rew & $\ms{7.34}{0.03}$ & $\ms{89.82}{0.29}$ & $\ms{49.93}{0.54}$ & $\ms{4.87}{0.05}$ & $\ms{7.19}{0.07}$ & $\ms{71.33}{0.67}$ \\
w/o recon & $\ms{7.33}{0.12}$ & $\ms{89.84}{0.37}$ & $\ms{50.12}{0.73}$ & $\ms{4.89}{0.08}$ & $\ms{7.16}{0.17}$ & $\ms{71.06}{1.58}$ \\
DreamReg & $\msb{7.16}{0.13}$ & $\msb{90.45}{0.14}$ & $\msb{51.87}{0.35}$ & $\msb{4.77}{0.07}$ & $\msb{7.08}{0.15}$ & $\msb{72.12}{1.62}$ \\
\bottomrule
\end{tabular}
\end{table}
We conduct ablations on CAMUS to evaluate the contribution of each supervision signal (Table~\ref{tab:ablation}). 
``only pose'' trains the policy using pose regression alone; 
``w/o rew'' removes the direction-based progress supervision; 
``w/o recon'' removes slice reconstruction supervision.
Removing the progress-alignment signal slightly degrades all metrics, 
% increasing DisErr and TransErr while reducing NCC, SSIM, and PNCC, 
indicating that reward-based progress supervision improves action quality and stabilizes belief updates. 
Removing reconstruction mainly decreases appearance consistency (SSIM and PNCC) and also increases geometric errors, suggesting that slice reconstruction regularizes latent dynamics. 
The pose-only variant performs worst overall, confirming that direct pose regression without belief-driven imagination is insufficient under partial observability. 
% Overall, the full DreamReg objective consistently achieves the best performance across both geometric and similarity metrics.

\section{Conclusion}
We proposed DreamReg, a belief-driven world-model framework for 2D--3D ultrasound registration. 
By modeling action–observation coupling, DreamReg performs iterative belief refinement instead of one-shot pose regression. 
Experiments on CAMUS and $\mu$-RegPro demonstrate consistent improvements in geometric and image similarity metrics under partial observability. 
The evaluation is conducted on 2D slices resampled from 3D volumes and does not model real tissue deformation, which may limit generalization to fully deformable settings. 
Although sequential rollout introduces additional computation, the achieved frame rate satisfies real-time clinical requirements. 
Future work will improve efficiency and extend the framework to deformable and closed-loop robotic ultrasound systems.
% We proposed DreamReg, a belief-driven world-model framework for 2D–3D ultrasound registration.
% By modeling the coupling between pose updates and subsequent observations, DreamReg enables iterative belief refinement rather than one-shot regression.
% Experiments on CAMUS and $\mu$-RegPro show consistent gains in geometric and image similarity metrics, demonstrating the robustness of action-conditioned latent dynamics under partial observability.
% Notably, our evaluation uses 2D slices resampled from 3D volumes as intra-operative observations, which neglects real-world tissue deformation and may limit generalization to fully deformable settings.
% Although sequential rollout introduces additional overhead compared with lightweight regression models, the achieved frame rate satisfies real-time clinical requirements.
% Future work will improve efficiency and extend the framework to deformable and closed-loop robotic ultrasound systems with adaptive, uncertainty-aware control.
% {\color{blue}Can you add some limitations of the work and future work to address these limitations. Computational time? dynamics/tissue deformation not considered? synthetic trajectories? etc. Actually, a dedicated paragraph on limitations would be good in section 3 (or you can put it under section 3.2 when you compare with SOTA methods).}
\begin{credits}
\subsubsection{\ackname} Research reported in this work was supported in part by Research Grants Council (RGC) of Hong Kong (CUHK 14217822, CUHK 14207823, CUHK 14211425, T45-401/22-N, and AoE/E-407/24-N) and in part by Innovation and Technology Commission of Hong Kong (MHP/096/22, ITS/235/22, ITS/224/23, ITS/225/23, and Multi-scale Medical Robotics Center (InnoHK initiative)).

\subsubsection{\discintname}
The authors have no competing interests to declare that
are relevant to the content of this article.
\end{credits}
\bibliographystyle{splncs04}
\bibliography{ref}
\end{document}